\documentclass[11pt]{article}

\usepackage[final]{acl}

\usepackage{times}
\usepackage{latexsym}

\usepackage[T1]{fontenc}

\usepackage[utf8]{inputenc}
\usepackage{amsfonts}
\usepackage{amsmath}
\usepackage{amssymb}
\usepackage{amsthm}
\usepackage{enumitem}
\usepackage{listings}
\usepackage{booktabs}
\usepackage{mathrsfs}
\usepackage[normalem]{ulem}
\definecolor{darkpink}{rgb}{0.96,0.14,0.56}
\definecolor{darkgreen}{rgb}{0,0.6,0.2} 

\usepackage{hyperref} 
 \hypersetup{ 
     colorlinks=true, 
     linkcolor=blue, 
     filecolor=blue, 
     urlcolor=darkpink, 
 } 
\usepackage{bbm}
\usepackage[ruled,vlined]{algorithm2e}

\usepackage{microtype}
\usepackage{multirow}

\usepackage{inconsolata}

\usepackage{graphicx}
\usepackage{CJKutf8}

\title{\fontsize{15}{16}\selectfont ProPy: Building Interactive Prompt Pyramids upon CLIP for Partially Relevant Video Retrieval}

\author{
  Yi Pan$^{1,2}$ \quad Yujia Zhang$^{1}$\thanks{Corresponding author.} \quad Michael Kampffmeyer$^{3}$ \quad Xiaoguang Zhao$^{1}$ 
  \\
  $^1$ State Key Laboratory of Multimodal Artificial Intelligence Systems, \\Institute of Automation, Chinese Academy of Sciences \\
  $^2$ School of Artificial Intelligence, University of Chinese Academy of Sciences \\
  $^3$ Department of Physics and Technology, UiT The Arctic University of Norway \\
  \{panyi2022,zhangyujia2014,xiaoguang.zhao\}@ia.ac.cn, michael.c.kampffmeyer@uit.no}

\begin{document}
\begin{CJK}{UTF8}{gbsn}
\maketitle
\begin{abstract}
Partially Relevant Video Retrieval (PRVR) is a practical yet challenging task that involves retrieving videos based on queries relevant to only specific segments.  While existing works follow the paradigm of developing models to process unimodal features, powerful pretrained vision-language models like CLIP remain underexplored in this field. To bridge this gap, we propose \textbf{ProPy}, a {model with systematic architectural adaption of CLIP specifically designed for PRVR}. Drawing insights from the semantic relevance of multi-granularity events, ProPy introduces two key innovations: (1) A \textbf{Prompt Pyramid} structure that organizes event prompts to capture semantics at multiple granularity levels, and (2) An \textbf{Ancestor-Descendant Interaction Mechanism} built on the pyramid that enables dynamic semantic interaction among events. With these designs, ProPy achieves SOTA performance on three public datasets, outperforming previous models by significant margins. Code is available at \href{https://github.com/BUAAPY/ProPy}{https://github.com/BUAAPY/ProPy}.
\end{abstract}

\section{Introduction}

Partially Relevant Video Retrieval (PRVR)~\cite{dong2022partially,wang2024gmmformer}  is a challenging task that retrieves videos  based on queries relevant to only specific segments. Unlike traditional Text-to-Video Retrieval (T2VR)~\cite{gabeur2020multi,luo2022clip4clip}, which requires the query to match the entire video, PRVR better aligns with real-world scenarios -- where long videos often consist of multiple events, and users may only be interested in specific segments. This makes PRVR more practical and promising for real-world applications.

\begin{figure}[t]
\centering
\includegraphics[width=1.0\columnwidth]{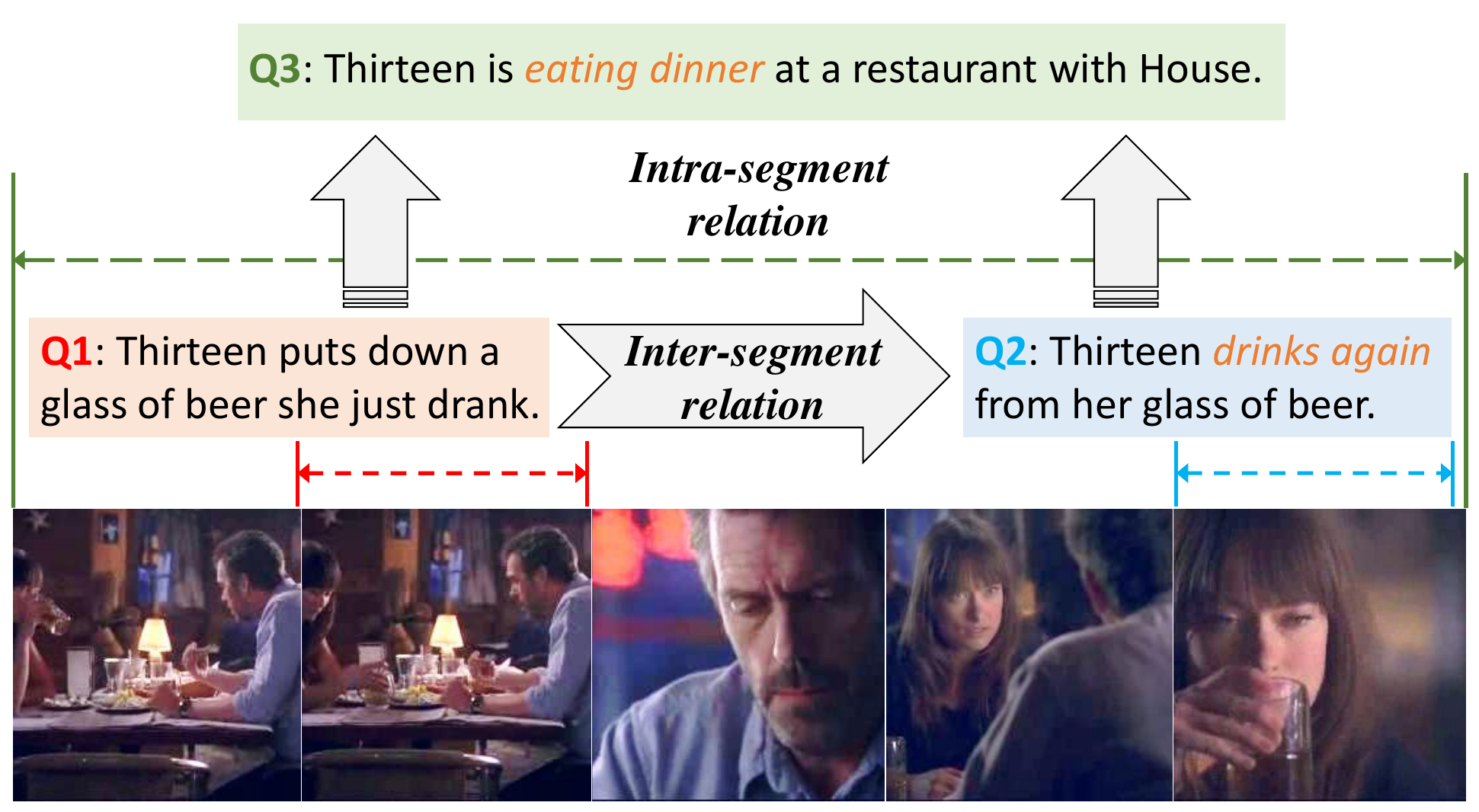}
\caption{Intra-segment relations and Inter-segment relations. The semantic understanding of Q2's `drink again' depends on contextual information from previous segments. Meanwhile, the high-level action `eating dinner' (relevant to Q3) is composed of lower-level, intra-segment events that correspond to Q1 and Q2.}
\label{fig_event}
\vspace{-0.6cm}
\end{figure}

Despite significant progress, most PRVR methods~\cite{dong2022partially,wang2024gmmformer,dong2023dual,jiang2023progressive,wang2024gmmformerv2,Li25_HLFormer} follow the paradigm of developing models to process extracted unimodal features. While pretrained vision-language models such as CLIP~\cite{radford2021learning} have shown remarkable success in T2VR~\cite{luo2022clip4clip}, their potential remains underexplored for PRVR. {A recent work QASIR~\cite{nishimura2023vision} introduces adapters on top of CLIP to process super-image features. However, this approach does not involve in-depth structural adaptions, leaving CLIP's capabilities not fully exploited. To  bridge this gap, we propose a model with systematic architectural adaption of CLIP specifically designed for PRVR.} Our approach builds upon recent prompt-based T2VR methods~\cite{yang2024dgl,zhang2024mpt,liu2025stop}, which demonstrate both effectiveness and efficiency by aggregating video semantics through an explicit global token. However, PRVR presents unique challenges: videos must be modeled as compositions of multiple events rather than encoded as single vector representations. A naive extension of T2VR approaches -- treating each segment as an independent sub-video -- fails to capture the rich semantic relationships between events.   Concretely, there are two fundamental types of event relationships that are crucial to model, as illustrated in Figure \ref{fig_event}: \textbf{intra-segment relations} representing compositional semantics between events with hierarchical inclusion relationships, and \textbf{inter-segment relations} capturing contextual dependencies between temporally distinct events. {The former ones are important in comprehending long events (\textit{eat dinner}) composed of multiple sub-events (\textit{drink}), while the latter ones are crucial in scenes where context semantics are required (\textit{drink again}).} Modeling these relations and ensuring semantic interactions of relevant events are beneficial for a comprehensive video understanding~\cite{fei2024video,yang2023towards}.

To effectively model the aforementioned event relations and their semantic interactions, we propose \textbf{ProPy} (Interactive \textbf{Pro}mpt \textbf{Py}ramid), a novel CLIP-based architecture for PRVR. ProPy leverages a set of event prompts focusing on segments of varying granularity and organizes them into a \textbf{Prompt Pyramid} based on the lengths and positions of their segments. This hierarchically structures a video into multi-granularity events. To account for the distinction between intra-segment and inter-segment relations, we design an \textbf{Ancestor-Descendant Interaction Mechanism}, which facilitates \textit{direct} interactions for intra-segment relations and \textit{indirect} interactions for inter-segment relations. Specifically, an ancestor-descendant relationship is established between two event prompts when their governed segments exhibit an inclusion relationship. Direct interactions are permitted only for intra-segment events, while inter-segment interactions are conducted indirectly through upper-level event prompts. With these carefully designed architectures and mechanisms, ProPy achieves SOTA performance on three challenging datasets, demonstrating the superiority of our method. Overall, our contributions can be summarized as follows:
\begin{itemize}[leftmargin=*]
    \item We propose ProPy, a novel solution to the PRVR task. To the best of our knowledge, ProPy is the first work that involves {systematic architectural designs} on pretrained vision-language models in the PRVR field.
    \item Based on the unique characteristics of PRVR, we design a Prompt Pyramid structure to process events with varying granularity, and an Ancestor-Descendant Interaction Mechanism to ensure sufficient semantic interactions for events with intra-segment and inter-segment relations .
    \item ProPy achieves SOTA performance with notable improvements on three public datasets, demonstrating its effectiveness and superiority.
\end{itemize}

\section{Related Work}
\noindent\textbf{Text-to-Video Retrieval} focuses on retrieving videos that fully match given textual queries~\cite{gabeur2020multi,luo2022clip4clip,yang2024dgl,huang2023vop}. Since the introduction of pretrained vision-language models~\cite{li2022blip,li2023blip2} like CLIP~\cite{radford2021learning} in the image domain, significant research efforts~\cite{jia2022visual,deng2023prompt,luo2022clip4clip,yang2024dgl,cao2024rap,liu2025stop} have been directed toward adapting these models for T2VR. Notably, recent prompt-based methods~\cite{zang2022unified,yang2024dgl,liu2025stop,zhang2024mpt,huang2023vop} have demonstrated competitive performance while maintaining efficiency through the use of only a small number of prompt tokens.

\noindent\textbf{Partially Relevant Video Retrieval} addresses the task of retrieving videos based on queries relevant to partial segments~\cite{dong2022partially,wang2024gmmformer}. Current PRVR approaches~\cite{dong2022partially,wang2024gmmformer,dong2023dual,cheng2024transferable,jun2025bridging,ren2025exploiting,Li25_HLFormer} predominantly adopt the Multiple Instance Learning (MIL) paradigm~\cite{waqas2024exploring}, employing coarse-fine two-branch architectures to model multiple events during both training and inference. While some recent works~\cite{song2025towards,moon2025prototypes} have incorporated pretrained vision-language models, they primarily utilize them for basic feature extraction without architectural innovations, or source for feature distillation~\cite{dong2023dual,zhang2025multi}. QASIR~\cite{nishimura2023vision}, the most similar work to ours, merely introduces adapters on top of CLIP to process super-image features while leaving the core CLIP layers unchanged. We argue that such surface-level modifications are insufficient to fully leverage CLIP's capabilities for the PRVR task.

\section{Methodology}
\begin{figure*}[!t]
\centering
\includegraphics[width=2.0\columnwidth]{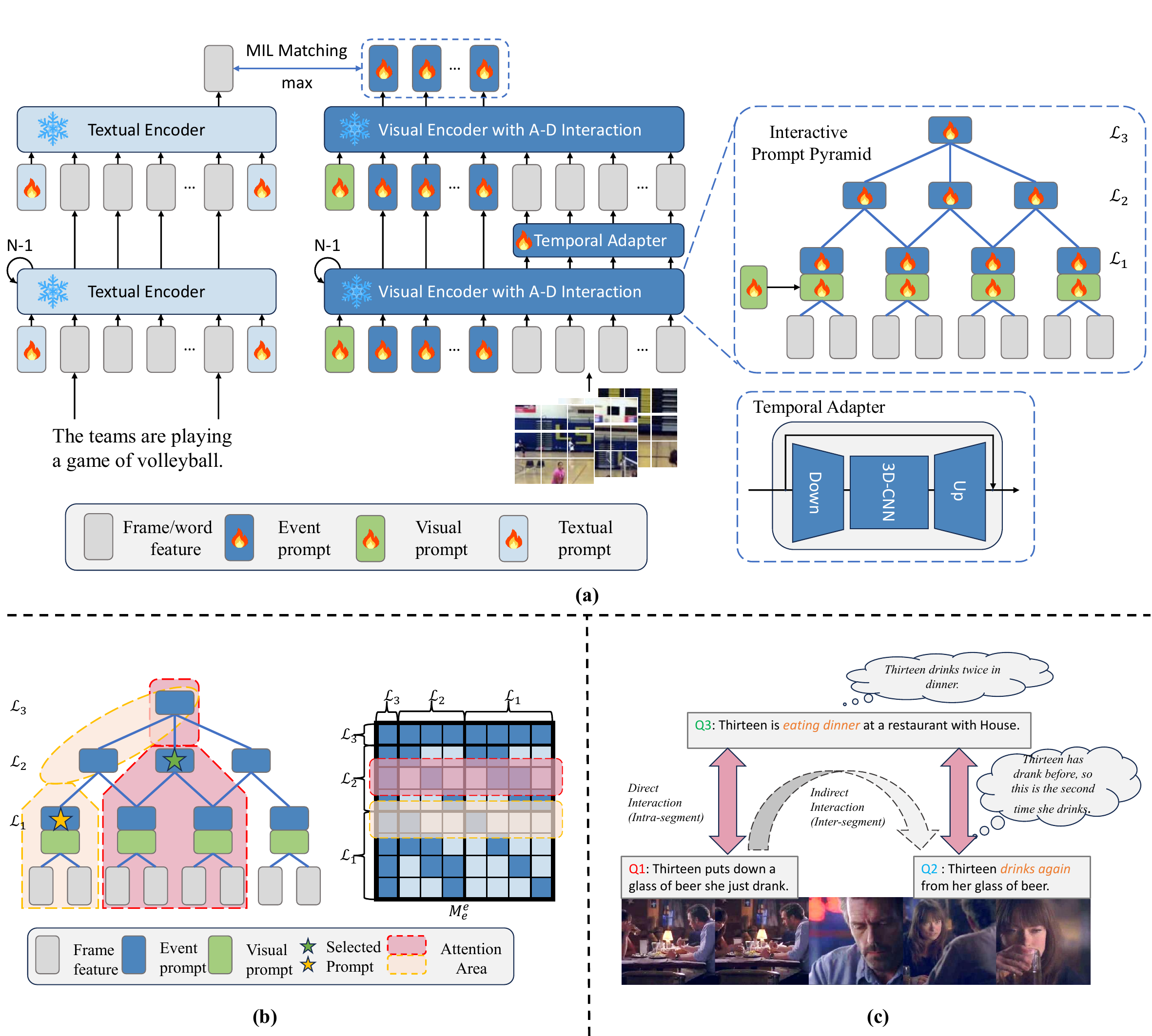}
\caption{\textbf{(a)}: Overview of ProPy. For the visual branch, the event Prompt Pyramid is built upon frame sequence and visual prompts, then event prompts are updated based on the Ancestor-Descendant (A-D) Interaction mechanism. Temporal Adapters are adopted for frame features to strengthen temporal semantics. For the textual branch, prefix and postfix textual prompts are added. We only show $8$ frames and a 3-layer pyramid for clarity. \textbf{(b)} Details of the Ancestor-Descendant Interaction Mechanism. \textbf{Left}: Attention areas of query event prompts. `selected prompt' means the event prompt served as queries during attention operation. \textbf{Right}: Attention mask $M^e_e$ of event prompts. Positions with attention scores are shown in dark blue. \textbf{(c)}: An example of direct interactions for intra-segment semantics and indirect interactions for inter-segment semantics.}
\label{fig_propy}
\end{figure*}

We formally define the PRVR task: Given a set of videos $\mathbb{V}=\{{V}_1,{V}_2,...{V}_{|\mathbb{V}|}\}$, each video ${V}_{i}$ can be represented as a list of $N_f$ frames: ${V}_{i}=\{f^i_1,f^i_2,...,f^i_{N_f}\}$. The PRVR task aims to retrieve videos with queries $T^i$ relevant only to certain segment $m^i_j$: ${V}_i=\underset{V\in \mathbb{V}}{\arg\max} P(V|T^i)$, where $m^i_j\subseteq {V}_i$ is a subset with consecutive frames.

\subsection{Overview of ProPy}
As shown in Figure~\ref{fig_propy} (a), ProPy deeply integrates with CLIP's visual and textual branches. The visual branch utilizes $N_e$ \textbf{event prompts} $E\in \mathbb{R}^{N_e \times d_v}$ (where $d_v$ is the ViT dimension) organized in a \textbf{Prompt Pyramid} to extract multi-granularity segment features.  For the $l$-th ViT layer, we add $N_v$ \textbf{visual prompts}~\cite{yang2024dgl} $P_{l}^{v}\in \mathbb{R}^{N_v\times d_v}$ and a \textbf{temporal adapter}~\cite{pan2022st}  $\Omega_l$ to extract spatial and temporal information. The \textbf{Ancestor-Descendant Interaction Mechanism} updates $E$ based on intra/inter-segment relations. The textual branch incorporates $N_t$ \textbf{text prompts} $P_l^t\in \mathbb{R}^{N_t\times d}$ (where $d$ is the dimension of CLIP) following DGL~\cite{yang2024dgl}. The model is trained based on Contrastive Learning~\cite{radford2021learning} and Multiple Instance Learning~\cite{waqas2024exploring} paradigms.

\subsection{Visual Branch}


\begin{table}[t]
\centering
\resizebox{\linewidth}{!}{\begin{tabular}{c|c}  
\hline                     
\textbf{Notation} & \textbf{Meaning} \\           
\hline                     
\textit{E} & set of all event prompts \\     
$\mathcal{L}_k$ & the $k$-th event prompt layer  \\
$e_j^k$ & the $j$-th event prompt from $\mathcal{L}_k$  \\
$m_j^k$ & the segment corresponding to $e_j^k$ \\
$n_k$ & number of event prompts from $\mathcal{L}_k$ \\
$c_k$ & the number of children of $e_j^k$\\
$o_k$ & the offset of children between $e_j^k$ and $e_{j+1}^k$\\
$\mathcal{A}(e_j^k)$ & the ancestors set of $e_j^k$ \\
$\mathcal{D}(e_j^k)$ & the descendants set of $e_j^k$ \\
$\mathcal{P}(e_j^k)$ & the parent set of $e_j^k$ \\
$\mathcal{C}(e_j^k)$ & the children set of $e_j^k$ \\

\hline                     
\end{tabular}}
\caption{Notations of prompt pyramid}  
\label{tab:notation}        
\end{table}

\noindent\textbf{Prompt Pyramid} 
We first detail the construction of the proposed Prompt Pyramid. Used notations are detailed in Table~\ref{tab:notation}. Given a video $V$ with $N_f$ frames, there are theoretically ${N_f\times(N_f+1)}/{2}$  segments with lengths ranging from $1$ to $N_f$. To save memory, we empirically set $N_f=2^K, K>1$, and select segments with lengths of $2^k (1\leq k\leq K)$. We then evenly and sparsely sample $n_k$ segments from those of length $2^k$, and pair them with hierarchically arranged learnable event prompts:
\begin{equation}
    E=\{\mathcal{L}_k=\{e^k_j | 1\leq j\leq n_k\}|1\leq k\leq K\}
\end{equation} where $\mathcal{L}_k$ is the $k$-th prompt layer containing $n_k$ event prompts sorted  by position.
$e^k_j$ corresponds to segment $m^k_j$ of length $2^k$. In total, there are $N_e=\sum_{k=1}^{K} n_k$ event prompts.

A prompt pair $(e^{k_1}_{j_1}, e^{k_2}_{j_2})$ forms an \textbf{Ancestor-Descendant} (A-D) relation if their governed segments satisfy an inclusion relationship, formally:
\begin{equation}
\begin{gathered}
    e^{k_1}_{j_1}\in \mathcal{A}(e^{k_2}_{j_2})\Leftrightarrow  e^{k_2}_{j_2}\in \mathcal{D}(e^{k_1}_{j_1}) \\ \Leftrightarrow  m^{k_2}_{j_2}\subsetneq m^{k_1}_{j_1}
\end{gathered}
\end{equation}
Specially, if $k_1= k_2+1$, then $(e^{k_1}_{j_1}, e^{k_2}_{j_2})$ forms a Parent-Child (P-C) relation, with corresponding sets denoted as $\mathcal{P}(e^{k_2}_{j_2})$ and $\mathcal{C}(e^{k_1}_{j_1})$. To construct a symmetrical pyramid, we set the number of children $c_k$ and the offset of their leftmost children $o_k$ as constants for prompts in the $k$-th layer, {formally}:
\begin{equation}\label{def_co}
    \begin{cases}
    c_{k}= |\mathcal{C}(e^{k}_{j_1}) |,1\leq j_1\leq n_k \\
    \mathscr{L}_{k}(j_1)= \underset{j_2}{\arg\min}\{e^{k-1}_{j_2} | e^{k-1}_{j_2}\in \mathcal{C}(e^{k}_{j_1})\} \\
    o_{k}= \mathscr{L}_{k}(j_1+1)- \mathscr{L}_{k}(j_1), 1\leq j_1<n_k
    \end{cases}
\end{equation}{where $\mathscr{L}_{k}(j_1)$ is an operation to find index of the leftmost child of $e^k_{j_1}$. $c_k$ and $o_k$ are subject to the following constraint:}
\begin{equation}
    \frac{n_k-c_{k+1}}{o_{k+1}}+1=n_{k+1}, o_{k+1} \mid (n_k-c_{k+1})   
\end{equation}\label{constraint}
This resembles the kernel-stride constraint in CNN~\cite{li2021survey}, treating $c_{k+1}$ as kernel size and $o_{k+1}$ as stride, but with two differences: 1) No padding is applied. 2) $o_{k+1}$ must divide $n_k-c_{k+1}$ exactly. For the top layer ($k=K$), the offset $o_K$ is set to $1$ ($n_{K-1}=c_{K}, n_{K}=1$). Given the frame count $N_f$, the prompt pyramid is uniquely determined once the hyperparameters $\mathcal{H}=\{(c_k,o_k)\}$ are specified. The impact of structure configuration is discussed in Sec~\ref{ana}.

\noindent\textbf{Ancestor-Descendant Interaction Mechanism}
Next, we describe the update mechanism for event prompts in the visual branch. Given a video $\textbf{V}$ with $N_f$ frames, CLIP first splits and embeds each frame into sequential features $F\in \mathcal{R}^{N_f\times N_s\times d_v}$, where $N_s$ denotes the sequence length (including the appended [CLS] token). These features are then processed by a ViT with $N$ layers. For the $l$-th ViT layer, three components participate in the updating process: the event prompts $E_l \in \mathbb{R}^{N_e \times d_v}$, the frame features $F_l\in \mathcal{R}^{N_f\times N_s\times d_v}$ and the per-layer visual prompts $P_l^v\in \mathbb{R}^{N_v\times d_v}$, which guide the spatial-temporal attention between event prompts and frame features.

We update $E_l$ using ViT's attention layer while keeping its weights frozen. For clarity, we first describe the update process (Figure~\ref{fig_propy} (b)) for a single event prompt $e_l^k$ (from the $k$-th layer; we omit the prompt index $j$ for simplicity), then generalize to parallel computation. The update consists of attention operations on three components: Firstly, $e_l^k$ attends to frames within its governed segments, denoted as $F_l(e_l^k)\in \mathbb{R}^{{2^k}\times N_s\times d_v}$, to produce segment features. Secondly, the prompt \textit{directly} interacts with its complete hierarchical context, including its ancestors, descendants and itself, denoted as $E_l(e^{k}_l)$:
\begin{equation}
    E_l(e^{k}_l)=\mathcal{A}(e^{k}_l)\cup \mathcal{D}(e^{k}_l) \cup \{e^{k}_l\}
\end{equation}
Thirdly, $e^{k}_l$ incorporates visual prompts $P_l^v$ to capture spatial-temporal semantics. To preserve the structure information, we replicate $P_l^v$  for $n_1$ times ($n_1$ is the number of event prompts from the bottom prompt layer $\mathcal{L}_1$), resulting in $\widetilde{P}_l^v\in \mathbb{R}^{n_1\times N_v\times  d_v}$. These augmented visual prompts correspond one-to-one with the bottom prompt layer $\mathcal{L}_1$. Given $e^{k}_l$, we first refer to its descendant prompts in $\mathcal{L}_1$ (or itself if $l=1$), then incorporate all corresponding visual prompts, denoted as $\widetilde{P}_l^v(e^{k}_l)$. In this way, for any prompt pairs with an Ancestor-Descendant relation, their visual prompts also exhibit an inclusion relation. These components serve as keys and values for the attention operation:
\begin{equation}
  \begin{cases}
      K/V(e^{k}_{l})=[F_l(m^k), E_l(e^{k}_{l}), \widetilde{P}_l^v(e^{k}_l)] 
      \\
      e^{k}_{l+1}=Attn(e^{k}_{l},K(e^{k}_{l}),V(e^{k}_{l}))  
    \end{cases}  
\end{equation} where $[\cdot,\cdot,\cdot]$ denotes the concatenation operation on the first dimension, and $F_l(m^k)$, $\widetilde{P}_l^v(e^{k}_l)$ are flattened into 2d tensor before concatenation. In practice, we use three attention masks to realize parallel computation, formally:

\begin{equation}
    \begin{cases}
    K_l =V_l = [F_l, E_l, \widetilde{P}_l^v] \\
    M=[M^e_f,M^e_e,M^e_v] \\
    E_{l+1}=Attn(E_l,K_l,V_l,mask=M)
    \end{cases}
\end{equation} where $M^e_f\in \mathbb{R}^{N_e\times (N_f\times N_s)}$, $M^e_e\in \mathbb{R}^{N_e\times N_e}$, $M^e_v\in\mathbb{R}^{N_e\times (n_1\times N_v)}$ are attention masks regarding frames, event prompts and visual prompts, respectively. A fast construction algorithm for these three masks is detailed in Appendix~\ref{mask_algo}.

Note that an event prompt can only access frames from its own segments, and other high-level semantics are exchanged only through other event prompts. This helps to prohibit feature leakage while preserving semantic interactions. Any two event prompts are guaranteed to share at least one common prompt (e.g, the global prompt from the top layer) for direct interaction, offering an indirect communication channel to exchange inter-segment semantics for them, as shown in Figure \ref{fig_propy} (c). This design ensures all event prompts are interconnected, with closely positioned events maintaining denser interaction pathways while distant events exhibit sparser connections, naturally mirroring both intra-segment and inter-segment relationships.

\noindent\textbf{Frame Feature Update} Previous T2VR works~\cite{yang2024dgl,zhang2024mpt} incorporate global prompts in frame-wise attention to capture temporal semantics. However, for PRVR, we find this approach ineffective, even underperforming models without frame feature updates (Section~\ref{ana}). We attribute this to the inherent uncertainty in MIL training that creates unstable information pathways. Instead, we adopt a more stable approach using adapters~\cite{pan2022st} $\Omega_l$ to mine temporal semantics directly from frame features, independent of event prompts. The temporal adapter $\Omega_l$ is composed of a down-projection, a 3d-CNN, and an up-projection. In detail, given the frame features $F_l\in \mathcal{R}^{N_f\times N_s\times d_v}$, where $N_f$ is the number of frames, and $N_s = H\times W +1$ is the length of flattened patch tokens with the [CLS] token, the temporal adapter only operates on patch tokens. Features are resized to 2d shapes before the CNN, then back to 1d sequences before the up-projection:
\begin{equation}
    \begin{cases}
    \widetilde{F}_l=F_l[:,1:,:]\in \mathcal{R}^{N_f\times (H\times W)\times d_v} \\
    F^{down}_{l} = \textit{Down}_l (\widetilde{F}_l) \in \mathcal{R}^{N_f\times H\times W\times (d_v//2)} \\ 
    F^{temp}_{l} = \textit{CNN}_{l}(F^{down}_{l})\in \mathcal{R}^{N_f\times H\times W\times (d_v//2)} \\
    F^{up}_{l} = \textit{Up}_{l}(F^{temp}_{l}) \in \mathcal{R}^{N_f\times (H\times W)\times d_v} \\ 
    F_{l+1}[:, 1:, :] = [F_{l}[:,0,:], F_{l}[:, 1:, :] + F^{up}_{l}] 
    \end{cases}
\end{equation}

The output event prompts from the last layer are projected to $d$ dimension to represent multi-granularity event features, denoted as $\widetilde{E}\in\mathbb{R}^{N_e\times d}$.

\subsection{Textual Branch}
The textual branch builds upon DGL~\cite{yang2024dgl}. To enhance multimodal alignment, two projection layers are utilized to project visual prompts $P_l^v$ to prefix and postfix prompts. These are concatenated with word features for updating:
\begin{equation}
\begin{cases}
        P_{l}^{pre/post}=f_{pre/post}(P_l^v)\in \mathbb{R}^{(Nt/2)\times d} \\
        [\_\_, T_{l+1}, \_\_] = L^{t}_{l}([P_{l,pre}^{t}, T_l, P_{l,post}^{t}]) 
\end{cases}
\end{equation}
where $L_l^t$ is the $l$-th layer of the textual branch, $P_{l}^{pre/post}$ are prefix and postfix prompts, $f_{pre/post}$ are projection layers, $T_l$ are input word features. The query representation $\widetilde{T}\in \mathbb{R}^{d}$ is obtained from the last word's features from the final layer.

\subsection{Training Objective}
Following the MIL paradigm, the highest similarity score between the query $\widetilde{T}$ and event prompts $\widetilde{E}$ is selected:
\begin{equation}
    S(T,V)=\underset{e}{max}\{cos(\widetilde{T},\widetilde{E})\}
\end{equation} The alignment is conducted with pair-wise similarities based on the symmetric InfoNCE loss~\cite{chen2020simple,radford2021learning}.

\section{Experiments}
\definecolor{darkgreen}{HTML}{00b050}
\definecolor{red}{HTML}{ff0000}
\definecolor{orange}{HTML}{faa702}
\definecolor{shallowblue}{HTML}{f0f7ff}
\definecolor{tsne_red}{HTML}{f41113}
\definecolor{tsne_blue}{HTML}{0501fa}

\begin{table*}[t]
\centering
\small
{ 
 \caption{Performance comparison on TVR, ActivityNet Captions and Charades-STA dataset. Rows highlighted in gray represent original performance of methods leveraging ResNet152 + I3D + Roberta features. The best, second and third performance are marked in \textbf{bold}, \underline{underline} and \uwave{wave} , respectively .}
\label{table_tvr_act_cha}
 \resizebox{1.0\linewidth}{!}{\begin{tabular}{l|ccccc|ccccc|ccccl} \hline
     \multicolumn{1}{c|}{\multirow{2}{*}{Method}} & \multicolumn{5}{c|}{TVR} & \multicolumn{5}{c|}{ActivityNet Captions} & \multicolumn{5}{c}{Charades-STA}\\  
    \multicolumn{1}{l|}{}  & R@1 & R@5 & R@10 & R@100 & SumR &  R@1 & R@5 & R@10 & R@100 & SumR  &  R@1 & R@5 & R@10 & R@100 & SumR \\ \hline

    \rowcolor{gray!20} MS-SL  &13.5 &32.1 &43.4 &83.4 &172.4 &7.1 &22.5 &34.7 &75.8 &140.1 &1.8 &7.1 &11.8 &47.7 &68.4\\ 
    \rowcolor{gray!20} PEAN &13.5 &32.8 &44.1 &83.9 &174.2 &7.4 &23.0 &35.5 &75.9 &141.8 &\textbf{2.7} &\uwave{8.1} &\uwave{13.5} &50.3 &\uwave{74.7} \\ 
    
    \rowcolor{gray!20} GMMFormer &13.9 &33.3 &44.5 &84.9 &176.6 &8.3 &24.9 &36.7 &76.1 &146.0 & 2.1 &7.8 &12.5 &\underline{50.6} &72.9\\ 
    \rowcolor{gray!20} DL-DKD &14.4 &34.9 &45.8 &84.9 &179.9 &8.0 &25.0 &37.5 &77.1 &147.6 &- &- &- &- &- \\
    \rowcolor{gray!20} Proto & 15.4  &35.9 &47.5 &86.4 & 185.1 &7.9 &24.9 &37.2 &77.3 &147.4 &- &- &- &- &-\\
    \rowcolor{gray!20} ARL &15.6 &36.3 &47.7 &86.3 &185.9 &8.3 &24.6 &37.4 &78.0 &148.3 &- &- &- &- &- \\
    \rowcolor{gray!20} GMMFormer-V2 &16.2 &37.6 &48.8 &86.4 &189.1 &8.9 &27.1 &40.2 &78.7 &154.9 &\uwave{2.5} &\underline{8.6} &\underline{13.9} &\textbf{53.2} &\textbf{78.2}\\ 
    
    \hline 
    DGL-MIL &17.5 &38.2 &51.1& 87.5 &194.3 & 10.5 & 26.4 & 40.5 & 77.4 &154.8 &1.4 &5.3 &9.2 &40.6 &56.5\\ 
    MS-SL &17.2 &39.1 &\uwave{51.5} &87.4 &195.2 &9.4 &26.1 &37.9 &77.2 &150.6 &1.3 &4.6 &8.2 &38.5 &52.6($\downarrow$15.8)\\ 
    QASIR &19.0 &39.9 &50.4 &87.2 &196.5 &\underline{14.1} &\underline{32.9} &\uwave{44.5} &\uwave{79.9} &\uwave{171.4} &1.9 &5.8 &10.1 &40.0 &57.8\\
    GMMFormer& 18.4 &39.5 &50.8 &\uwave{89.2} &197.9 &9.4 &26.4 &38.2 &76.2 &150.2 &0.9 & 4.5 & 8.0 & 39.0 & 52.4($\downarrow$20.5)\\
    GMMFormer-V2 &\uwave{19.2} &\uwave{40.1} &50.5 &\textbf{90.3} &\uwave{200.1} &10.8 &28.8 &41.1 &78.1 & 158.8 & 1.3 & 4.7 & 9.0 & 40.4 & 55.4($\downarrow$22.8)\\

    AMDNet &\underline{19.7} &\underline{42.4} &\underline{54.1} &88.9 &\underline{205.1} &\uwave{12.3} &\uwave{32.5} &\underline{45.9} &\underline{82.1} &\underline{172.8} &1.1 &4.2 &7.2 &36.4 &48.9\\
    \hline 
    \rowcolor{shallowblue}Propy & \textbf{22.4} &\textbf{45.0} &\textbf{55.9} &\underline{89.5} &\textbf{212.8} &\textbf{14.9} &\textbf{34.9} &\textbf{47.5} & \textbf{82.7} &\textbf{180.0}  &\underline{2.6} &\textbf{8.7} &\textbf{14.8} &\uwave{50.4} &\underline{76.5} \\ 
\hline 
 \end{tabular}}
}
\vspace{-0.2cm}
\end{table*}

\begin{table}[t]
{
 \centering
 \small
 \caption{Performance on QVHighlights $val$ split. Rows highlighted in gray are results with CLIP-B/16 features adopted from Proto~\cite{moon2025prototypes}.
 }
\label{table_qv}
 \resizebox{\linewidth}{!}{\begin{tabular}{l|ccccc}
 \hline
Model  & R@1 & R@5 & R@10 & R@100 & SumR  \\ \hline
\rowcolor{gray!20} GMMFormer &18.2 &43.7 &56.7 &92.5 &211.1 \\
\rowcolor{gray!20} MS-SL &20.4 &46.7 &60.7 &\underline{94.6} &222.5 \\
\rowcolor{gray!20} Proto &\underline{22.6} &\underline{48.8} &\underline{61.3} &{93.9} &\underline{226.6} \\
\hline 
GMMFormer &16.3 &39.7 &52.3 &88.4 &196.7\\
AMDNet &17.1 &40.8 &52.5 &88.4 &198.8\\
GMMFormer-V2 &15.6 &40.2 &53.7 &88.5 &198.0\\
MS-SL &17.4 &43.4 &55.2 &88.8 &204.8\\ 
\hline 
\rowcolor{shallowblue}Propy & \textbf{37.4} &\textbf{65.6} &\textbf{76.1} &\textbf{96.5} &\textbf{275.5} \\ \hline
 \end{tabular}}
}
\vspace{-0.2cm}

\end{table}

\begin{table}[t]
{
 \centering
 \small
 \caption{Ablation on pyramid structures. 
 }
\label{table_structure}
 \resizebox{\linewidth}{!}{\begin{tabular}{c|c|ccc}
 \hline
$N_f$ & $\mathcal{H}=\{(c_k,o_k)\}$ & R@5 & R@10 & R@100  \\ \hline
16 & \{(4,2),(3,2),(3,1)\}  &8.3 &14.1 &47.7\\
16 & \{(2,1),(3,2),(3,2),(3,1)\}  &8.5 &14.2 &48.1\\
32 & \{(4,2),(3,2),(3,2),(3,1)\} &8.5 &14.4 &49.3\\
\hline 
\rowcolor{shallowblue}32 & \{(2,2),(2,1),(3,2),(3,2),(3,1)\} &\textbf{8.7} &\textbf{14.8} &\textbf{50.4}  \\ \hline
 \end{tabular}}
} 
\vspace{-0.2cm}
\end{table}

\begin{table}[t]
{
 \centering
 \small
 \caption{Ablation study on semantic interaction mechanism of event prompts.
 }
\label{table_sem_inter}
 \resizebox{\linewidth}{!}{\begin{tabular}{c|c|ccc}
 \hline
Attention Area & Interaction & R@5 & R@10 & R@100   \\ \hline
$\mathcal{A}$& inter-only  &8.1 &12.6 &49.5 \\
$\mathcal{D}$& intra-only  &8.4 &13.4 &48.3 \\
$\mathcal{P}$& inter-only &8.5 &13.7 &49.1  \\
$\mathcal{C}$& intra-only &8.3 &13.3 &47.8 \\
$\mathcal{P}\cup \mathcal{C}$&inter-intra  &8.6 &14.1 &49.4 \\
$\mathcal{W}$& unstructured  &\textbf{8.7} &14.2 &48.7 \\
$\mathcal{S}$& none  &8.3 &14.1 &48.6 \\
\hline 
\rowcolor{shallowblue}$\mathcal{A}\cup \mathcal{D}$ & inter-intra &\textbf{8.7} &\textbf{14.8} &\textbf{50.4}  \\ \hline
 \end{tabular}}
}  

\vspace{-0.2cm}
\end{table}

\subsection{Experimental Settings}
\noindent\textbf{Datasets}
We evaluate ProPy on four public datasets: TVR~\cite{lei2020tvr}, ActivityNet-Captions~\cite{krishna2017dense}, Charades-STA~\cite{gao2017tall} and QVHighlights~\cite{lei2021detecting}. TVR comprises around 21.8K videos, each paired with 5 descriptions. ActivityNet-Captions (referred to as ActivityNet for simplicity) comprises around 20K YouTube videos, with an average duration of 118 seconds and 3.7 descriptions per video. Charades-STA contains 6.7K videos, annotated with an average of 2.4 descriptions per video.  QVHighlights is a dataset for moment retrieval containing over 10K videos. We follow previous works~\cite{dong2022partially,moon2025prototypes} for data partitioning and evaluation metrics.

\noindent\textbf{Implementation Details}
We select CLIP-B/32 as the backbone. The dimensions $d_v$, $d$ are set to 768 and 512, respectively. For the default structure hyperparameter, $N_f$ is set to $32$ and $\mathcal{H}$ is configured as $\{(2,2),(2,1),(3,2),(3,2),(3,1)\}$. Following DGL~\cite{yang2024dgl}, we use 4 visual prompts ($N_v$) and 8 textual prompts ($N_t$) per layer. The learning rates are set to 1e-3 for ActivityNet, 9e-4 for QVHighlights, and 8e-4 for TVR and Charades-STA, with a uniform batch size of 24 across all datasets. Following ~\cite{luo2022clip4clip,yang2024dgl}, ProPy is trained for 10 epochs using the AdamW optimizer. All experiments are conducted on a single NVIDIA RTX 3090 GPU.

\noindent\textbf{Baselines}
ProPy is the first prompt-based model built on CLIP for PRVR, and no existing methods adopt a similar architecture. To  ensure a comprehensive comparison, we evaluate 3 types of baseline models. (1) \textbf{DGL-MIL} We adapt our base framework, DGL~\cite{yang2024dgl}, to fit the MIL paradigm. Specifically, we expand the number of global prompts to $N_e$ and use the entire set for alignment.  (2) \textbf{QASIR}~\cite{nishimura2023vision}. A CLIP-based model that employs super-image construction for feature enhancement. (3) \textbf{Other PRVR models}~\cite{dong2022partially, dong2023dual, wang2024gmmformerv2, wang2024gmmformer, jiang2023progressive, song2025towards,  moon2025prototypes, cho2025ambiguity}. We run prior open-source PRVR models using extracted CLIP-B/32 features for fair comparison. Additionally, we include their original performance (trained with ResNet152~\cite{he2016deep}+I3D~\cite{carreira2017quo}+RoBERTa~\cite{liu2019roberta} features) as a reference. Further implementation details are provided in Appendix~\ref{ap_details}.

\subsection{Overall Comparison}
The performance comparison is shown in Table \ref{table_tvr_act_cha}, \ref{table_qv}.  As shown in Table \ref{table_tvr_act_cha}, ProPy significantly outperforms all baselines on TVR and ActivityNet, achieving absolute improvents of $7.7\%$ on TVR and $7.2\%$ on ActivityNet. These gains can be partially attributed to CLIP's well-aligned multimodal features, as evidenced by the improved performance of other PRVR models when using CLIP features. However, on Charades-STA, most PRVR models exhibit degraded performance with CLIP features, while ProPy remains competitive. This aligns with observations in prior work~\cite{nishimura2023vision}, where shorter query lengths in Charades-STA (averaging 6.2 words, vs. 12.2 for ActivityNet and 13.6 for TVR) lead to insufficient textual supervision~\cite{moon2025prototypes}, demanding deeper video understanding. In this circumstance, CLIP's static image features (from its last layer) underperform dynamic I3D features from 3D-CNNs. ProPy addresses this limitation by aggregating semantically rich features from \textit{all} CLIP layers and integrated temporal adapters, enabling comprehensive video understanding. On the challenging QVHighlights dataset, ProPy achieves remarkable gains (Table \ref{table_qv}) -- even surpassing methods with CLIP-B/16 (a more powerful backbone with 16x16 grid size) features -- with a $37.4\%$ R@1 score, far exceeding competitors. These results validate ProPy's superiority and broad applicability.

\subsection{Analysis}\label{ana}
Unless otherwise specified, the following experiments are based on the Charades-STA dataset.

\noindent\textbf{Pyramid Structure Settings}  We investigate the impact of \textit{width} ($N_f$) and \textit{depth} (number of layers), as shown in Table \ref{table_structure}. Results indicate that increasing either parameter improves performance by providing richer information and more event candidates. However, the memory issue should also be considered in model design, particularly for frame feature memory relevant to parameter $N_f$.

\noindent\textbf{Ancestor-Descendant Interaction Mechanism} We analyze the impact of interaction mechanisms. We evaluate 7 alternative mechanisms: 1) $\mathcal{A}$ attends only to ancestors, missing descendant intra-segment information; 2) $\mathcal{D}$ attends only to descendants, blocking inter-segment communication guided by ancestors; 3) $\mathcal{P}$ attends only to parent nodes, providing limited inter-segment interaction; 4) $\mathcal{C}$ attends only to child nodes, offering partial intra-segment interaction; 5) $\mathcal{P}\cup\mathcal{C}$ combines both parent and child attention; 6)  $\mathcal{W}$ attends to all nodes without structure; and 7) $\mathcal{S}$ prohibits any interaction.  The results in Table \ref{table_sem_inter} demonstrate that: (1) Semantic interactions are essential, as the non-interactive model $\mathcal{S}$ performs poorly; (2) Interaction structure significantly impacts performance, shown by $\mathcal{W}$'s results; (3) Both intra-segment and inter-segment interactions are important ($\mathcal{A}\cup\mathcal{D}$ is better than $\mathcal{A}$ and $\mathcal{D}$, $\mathcal{P}\cup\mathcal{C}$ is better than $\mathcal{P}$ and $\mathcal{C}$). (4) The Ancestor-Descendant Interaction Mechanism achieves optimal performance by effectively integrating both interaction types.

\begin{figure*}[t]
\centering
\includegraphics[width=2.0\columnwidth]{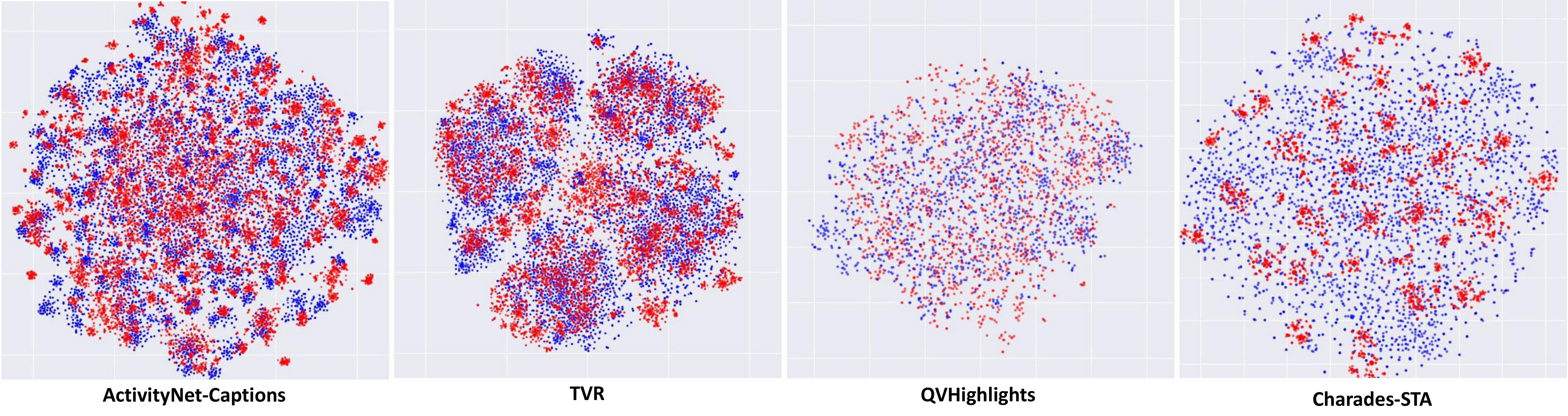}
\caption{TSNE visualizations. Points in \textcolor{tsne_red}{red}, \textcolor{tsne_blue}{blue} are text features and segment features, respectively.}
\label{fig_tsne} 
\vspace{-0.2cm}
\end{figure*}

\begin{figure*}[t]
\centering
\includegraphics[width=2.0\columnwidth]{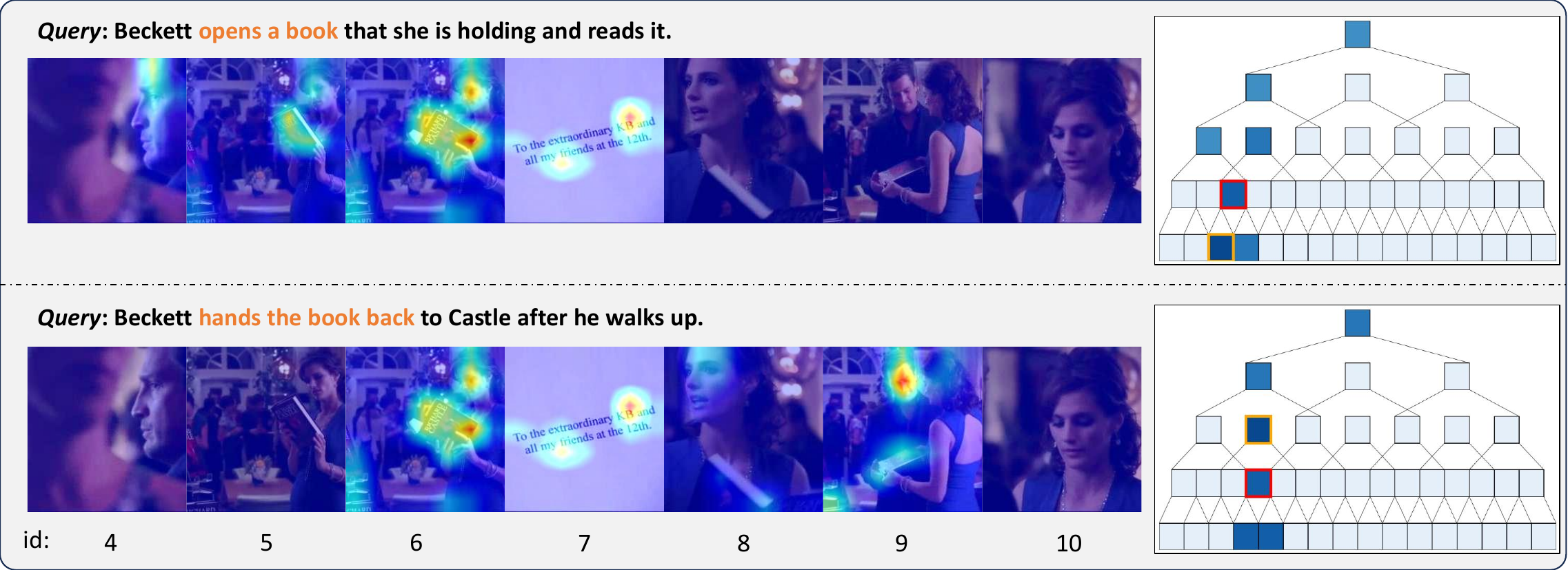}
\caption{\textbf{Left}: Retrieval results of ProPy (samples are selected based on the R@1 metric). The attention maps visualize the scores between selected event prompts and image patches. \textbf{Right}: Attention scores between selected event prompts (with \textcolor{red}{red} borders) and other prompts. Events with the highest scores are marked by \textcolor{orange}{orange} borders.}
\label{fig_res} 
\vspace{-0.2cm}
\end{figure*}

\begin{table}[t]
{
 \centering
 \small
 \caption{Ablation study on frame feature updating mechanism. 
 }
\label{table_frame_updating}
 \resizebox{\linewidth}{!}{\begin{tabular}{c|ccccc}
 \hline
Mechanism  & R@1 & R@5 & R@10 & R@100 & SumR  \\ \hline
attn-pyr & 1.5 &5.9 &9.5 &37.1 &54.0 \\
attn-whole & 1.4 &6.4 &10.4 &37.3 &55.5 \\
attn-adapter & 1.7 &6.6 &9.8 &40.0 &58.1 \\
orig & 1.9 &6.6 &11.3 &42.5 &62.3 \\
\hline 
\rowcolor{shallowblue}adapter & \textbf{2.6} &\textbf{8.7} &\textbf{14.8} &\textbf{50.4} &\textbf{76.5} \\ \hline
 \end{tabular}}
} 
\vspace{-0.2cm}
\end{table} 

\begin{table}[t]
{
 \centering
 \small
 \caption{Ablation study on event prompts from different layers for MIL learning.
 }
\label{table_levels}
 \resizebox{\linewidth}{!}{\begin{tabular}{c|ccccc}
 \hline
Levels(k)  & R@1 & R@5 & R@10 & R@100 & SumR  \\ \hline
\{1,2,3\} &2.4 &8.1 &13.0 &48.8 &72.3\\
\{3,4,5\} &2.3 &8.0 &13.2 &48.5 &72.0\\
\{1,2,3,4\} &\textbf{2.6} &\textbf{9.0} &14.3 &49.9 &75.8\\
\{2,3,4,5\} &2.3 &8.3 &13.6 &49.1 &73.3\\
\hline 
\rowcolor{shallowblue}\{1,2,3,4,5\} & \textbf{2.6} &8.7 &\textbf{14.8} &\textbf{50.4} &\textbf{76.5} \\ \hline
 \end{tabular}}
} 
\vspace{-0.2cm}
\end{table}

\begin{table}[t]
{
 \centering
 \small
 \caption{Ablation study of operations on visual prompts. 
 }
\label{table_visual_prompts}
 \resizebox{\linewidth}{!}{\begin{tabular}{c|ccccc}
 \hline
Operation  & R@1 & R@5 & R@10 & R@100 & SumR  \\ \hline
no-copy &2.2 &8.2 &13.0 &48.8 &72.2\\
$C(E)$ &2.1 &8.1 &13.3 &48.5 & 72.0\\
\hline 
\rowcolor{shallowblue}$C(\mathcal{L}_1)$ & \textbf{2.6} &\textbf{8.7} &\textbf{14.8} &\textbf{50.4} &\textbf{76.5} \\ \hline
 \end{tabular}}
} 
\end{table}

\begin{table}[t]
{
 \centering
 \small
 \caption{Ablation on each design. $\mathcal{A}\mathcal{D}$, $\Omega$, $C(\mathcal{L}_1)$ mean Ancestor-Descendant Interaction, using adapters and replicating visual prompts $n_1$ times, respectively. Compared operations are disabling event prompt interaction, vanilla CLIP processing and no replication.
 }
\label{table_comp}
 \resizebox{\linewidth}{!}{\begin{tabular}{c|ccc|ccccc}
 \hline
Idx  & $\mathcal{A}\mathcal{D}$ & $\Omega$ & $C(\mathcal{L}_1)$ & R@1 & R@5 & R@10 & R@100 & SumR \\ \hline
a &  &  &  &1.8 &5.8 &9.3 & 38.6 &55.5 \\ \hline 
b & \checkmark &  &  &1.7 & 6.8 &11.0 & 41.9 &61.4 \\
c &  & \checkmark &  &1.5 & 6.6 &10.9 &41.5 &60.5\\
d &  &  &\checkmark &1.7 &5.6 &9.4 &39.4 &56.1\\ 
e & \checkmark & \checkmark &  & 2.2 & 8.2 & 13.0 & 48.8 & 72.2 \\
f & \checkmark &  &\checkmark & 1.9 &6.6 &11.3 &42.5 &62.3\\ 
g &  & \checkmark & \checkmark & 2.1 &8.3 &14.1 &48.6 &73.1 \\
\hline 
\rowcolor{shallowblue}h & \checkmark & \checkmark &\checkmark & \textbf{2.6} &\textbf{8.7} &\textbf{14.8} &\textbf{50.4} &\textbf{76.5} \\ \hline
 \end{tabular}}
}
\vspace{-0.2cm}
\end{table}

\begin{table}[h]
\centering
\small
{ 
 \caption{Performance on the ActivityNet dataset in the Weakly-Supervised VCMR setting. Rows highlighted in gray represent original performance of methods leveraging ResNet152 + I3D + Roberta features. }
\label{table_vcmr_act}
 \resizebox{\linewidth}{!}{\begin{tabular}{l|cc|cc|cc} \hline
 \multicolumn{1}{c|}{\multirow{2}{*}{Method}} & \multicolumn{2}{c|}{IoU=0.3} & \multicolumn{2}{c|}{IoU=0.5}  & \multicolumn{2}{c}{IoU=0.7} \\  
\multicolumn{1}{l|}{}  & R@10 & R@100 & R@10 & R@100 & R@10 & R@100  \\ \hline
\rowcolor{gray!20}FAWL  &11.86 &38.98 &6.25 &21.77 &2.88 &10.05\\
\rowcolor{gray!20}JSG &13.27 &40.61 &8.76 &29.98 &3.83 &15.78\\
\hline 
FAWL &23.68 &48.02 &17.54 &43.57 &9.35 &20.66\\

JSG &25.62 &54.31 &19.35 & 45.15 & 10.92 & 26.14\\
\hline 
\rowcolor{shallowblue}Propy & \textbf{28.57} & \textbf{57.42} &\textbf{20.81} &\textbf{46.22} &\textbf{12.94} &\textbf{31.85}\\ 
\hline 
 \end{tabular}}
} 
\end{table}

\begin{figure}[t]
\centering
\includegraphics[width=1.0\columnwidth]{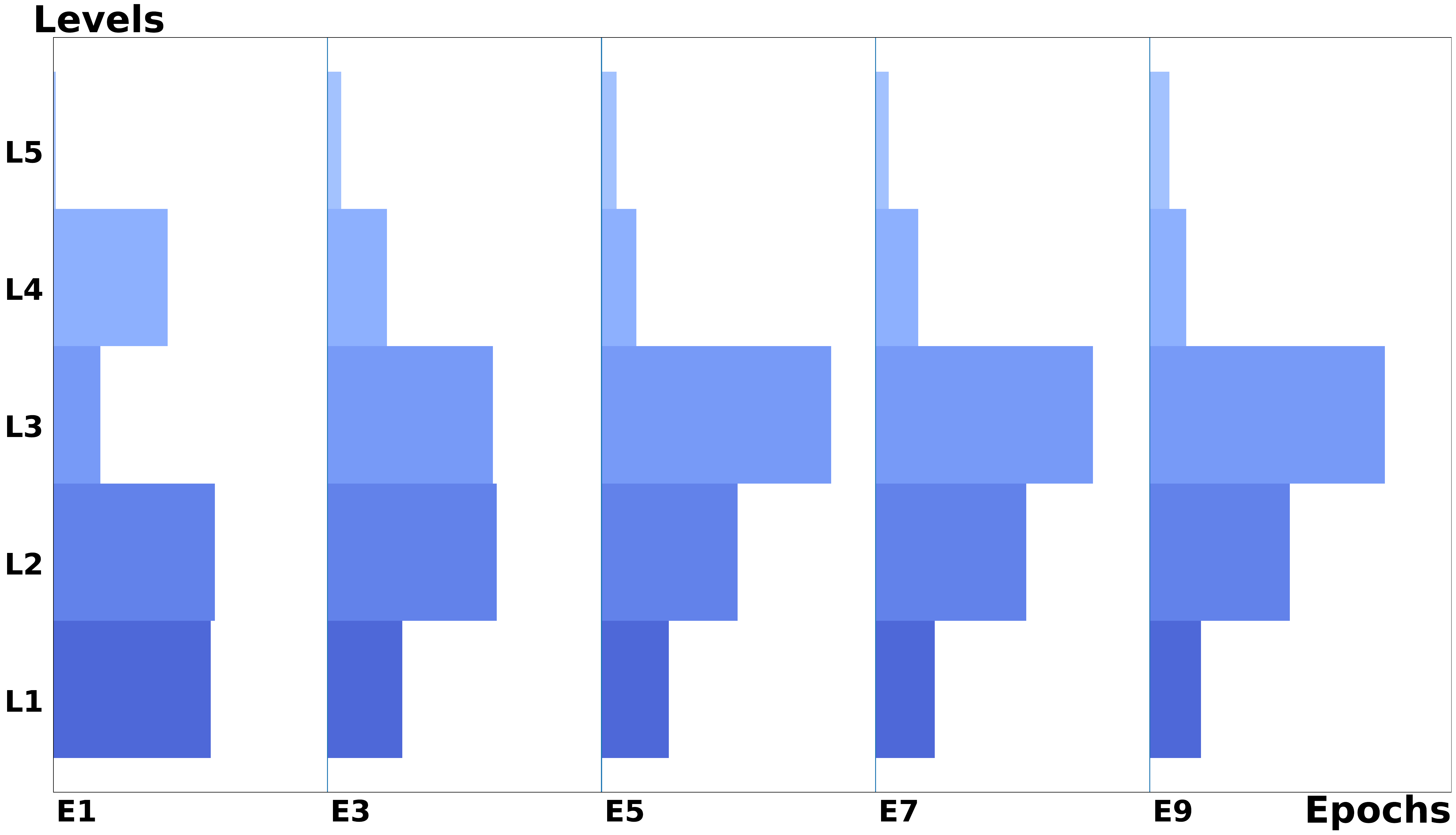}
\caption{Length distributions of selected events.}
\label{fig_dis} 
\vspace{-0.2cm}
\end{figure}

\begin{table}[h]
{
 \centering
 \small
 \caption{Parameters and inference time (per query) comparison on Charadest-STA. Matching time refers to the time for video-text matching process.
 }
\label{table_eff}
 \resizebox{\linewidth}{!}{\begin{tabular}{c|cc|cc}
 \hline
 \multicolumn{1}{c|}{\multirow{2}{*}{Method}} & \multicolumn{2}{c|}{Parameters(MB)} & \multicolumn{2}{c}{Inference Time(ms)} \\  
 \multicolumn{1}{l|}{} & trainable & total & matching &total\\ \hline 
MS-SL & 4.85 &4.85 & 0.65 &3.05 \\
GMMFormer & 12.85 &12.85  & 0.44 &1.79  \\
\hline 
\rowcolor{shallowblue}ProPy & 7.97 &159.24 & 1.16 &20.3\\
\hline 
 \end{tabular}}
}
\vspace{-0.2cm}
\end{table}

\noindent\textbf{Updating Mechanism of Frame Feature} We perform studies for the frame feature updating mechanism, evaluating 3 other distinct mechanisms: (1) \textit{orig}: vanilla CLIP processing without trainable components; (2) \textit{attn-whole}: DGL-style attention using \textit{all} event prompts as keys/values; (3) \textit{attn-pyr}: constrained attention where frames only interact with governing event prompts (masked by $M^e_f$);  (4) \textit{attn-adapter}: enhanced version of (3) with per-layer adapters. As shown in Table \ref{table_frame_updating}, the significant performance improvement demonstrates the effectiveness of temporal adapters.

\noindent\textbf{Operation on Visual Prompt} To validate the operation on visual prompts, we further consider two alternative operations: (1) \textit{no-copy}. All event prompts incorporate $N_v$ visual prompts for updating. The corresponding attention mask $M^e_v$ has the shape of $N_e\times N_v$. (2) $C(E)$. Visual prompts are copied $N_e$ times and assigned to event prompts in a one-to-one manner. $M^e_v$ is in the shape of $N_e\times (N_e\times N_v)$. The original design is denoted as $C(\mathcal{L}_1)$, which copies visual prompts $n_1$ times. Results are shown in Table \ref{table_visual_prompts}, showing that the  $C(\mathcal{L}_1)$ design achieves the best performance. This design considers the inclusion relation of event prompts, preserving beneficial structure information.

\noindent\textbf{Different Levels of Event Prompts} We select different levels of event prompts for MIL learning with the pyramid structure fixed to examine the roles of different prompt layers. Results in Table \ref{table_levels} show that models with more levels of prompts in learning are superior to those with fewer levels. We also find that layers nearer to the bottom (k=1) contribute more to performance. To uncover the underlying reason, we conduct a statistical experiment on the distribution of selected events during learning, shown in Figure~\ref{fig_dis}. It illustrates that ProPy learns in an easy-to-hard manner: it initially processes shorter segments then progressively extends to longer ones, with learned semantics evolving from low-level to high-level.

\noindent\textbf{Ablation Study} We conduct ablation studies on each design. As shown in Table~\ref{table_comp}, the Ancestor-Descendant Mechanism individually contributes the most. When equipped with adapters, the model achieves comparable performance with the Ancestor-Descendant Mechanism or the $C(\mathcal{L}_1)$ operation (e and g). This indicates that structured visual prompts are beneficial for event learning. The full setting, with temporal adapters, Ancestor-Descendant Mechanism and structured visual prompts, achieves the best performance. 

\noindent\textbf{Grounding Capability} We evaluate ProPy's grounding capability under the Weakly-Supervised Video Corpus Moment Retrieval~\cite{chen2023joint} setting. As shown in Table \ref{table_vcmr_act}, ProPy achieves SOTA performance on AcivityNet thanks to the design of the multi-granularity event pyramid. Notably, compared to previous methods~\cite{chen2023joint,pan2025fawl}, ProPy does not require complex intra-video losses and time-consuming NMS~\cite{lin2018bsn} operations.

\noindent\textbf{Efficiency Comparison} As shown in Table \ref{table_eff}, ProPy contains relatively more parameters. However, most parameters are frozen CLIP weights, and there are only $5\%$ trainable parameters. Among these trainable parameters, temporal adapters introduce $7.11$M ($89\%$), and prompts only occupy $0.86$M ($11\%$). However, as discussed before, incorporating these temporal adapters is crucial for overall performance. ProPy takes longer inference time, yet most time is spent on feature computation through CLIP layers. Once the computation process of features is done, the matching time is comparable with other models. This indicates that in real-world retrieval scenarios, ProPy enjoys high retrieval accuracy without significant increases in latency.

\noindent\textbf{Qualitative Analysis}
We  visualize the t-SNE clustering results in Figure~\ref{fig_tsne}. As observed, the cluster distributions vary across datasets, reflecting the differences in their video content and textual annotations. For ActivityNet, which primarily focuses on actions or events, semantically similar actions are located close to each other, resulting in a large number of distinct cluster centers. In TVR, the videos are sourced from 6 different TV shows, where each show features recurring characters and scenes, leading to roughly 4 to 6 clusters. QVHighlights contains videos from news and vlogs, with more diverse visual content and textual descriptions, which results in a more diffuse distribution without clear cluster centers. As previously discussed, the Charades-STA dataset contains short textual queries with limited fine-grained annotations, causing its textual features to collapse around a few dense centers. Except for the Charades dataset, the t-SNE plots of the other datasets generally reveal a reasonable degree of alignment between the video and textual features in the shared feature space.

We further illustrate some retrieval results from TVR in Figure~\ref{fig_res}. The results show that: (1) ProPy is able to extract high-quality spatial-temporal semantics, such as`opens a book' and `hands the book back'. (2) ProPy ensures sufficient semantic interactions. For example, the latter event `hands the book back to Castel' requires the previous context of `book', `Castel', and the visualization shows that the selected event gives the highest attention to its parent, which is one of direct information channels for the former event.

\section{Conclusion}
We propose ProPy, the first in-depth CLIP-based model for PRVR. By considering both intra-event and inter-event relations of video events, we design an Interactive Prompt Pyramid architecture to extract multi-granularity event features and an Ancestor-Descendant Interaction Mechanism to ensure sufficient semantic interactions. Extensive experiments demonstrate the superiority and generalizability of ProPy.
\section{Limitations}

Though ProPy only requires a small number of trainable parameters, the memory occupied by CLIP features cannot be ignored. Furthermore, the $2^k$ segment sampling strategy and the structure parameters $\mathcal{H}$ are empirical. Future works will include an adaptive selection method (like~\cite{wang2024videotree}) of video frames and pyramid structures.
\section*{Acknowledgement} 
This work was supported by the International Partnership Program of the Chinese Academy of Sciences (Grant No. 104GJHZ2023053FN) and the Young Scientists Fund of the State Key Laboratory of Multimodal Artificial Intelligence Systems (Grant No. ES2P100118).

\bibliography{main}

\appendix
\clearpage
\section{More Implementation Details}\label{ap_details}
Following~\cite{yang2024dgl}, videos are compressed first to 3fps and $224\times 224$ resolution. The model is trained using the AdamW optimizer whose decoupled weight decay is set to 0.2. During training, a warm-up strategy is adopted followed by a cosine learning rate policy. For other PRVR models, we utilize the output sequences from the last layer of CLIP's textual branch as text features, and the [CLS] features from the visual branch's last layer as visual features. To make minimal modifications,  the number of frames is set to $128$ for other PRVR models, which is much larger than $32$ for ProPy. The training process for baseline PRVR models also follows the original process, i.e., 100 max epochs with an early stop strategy.

\begin{figure}[h]
\centering
\includegraphics[width=1.0\columnwidth]{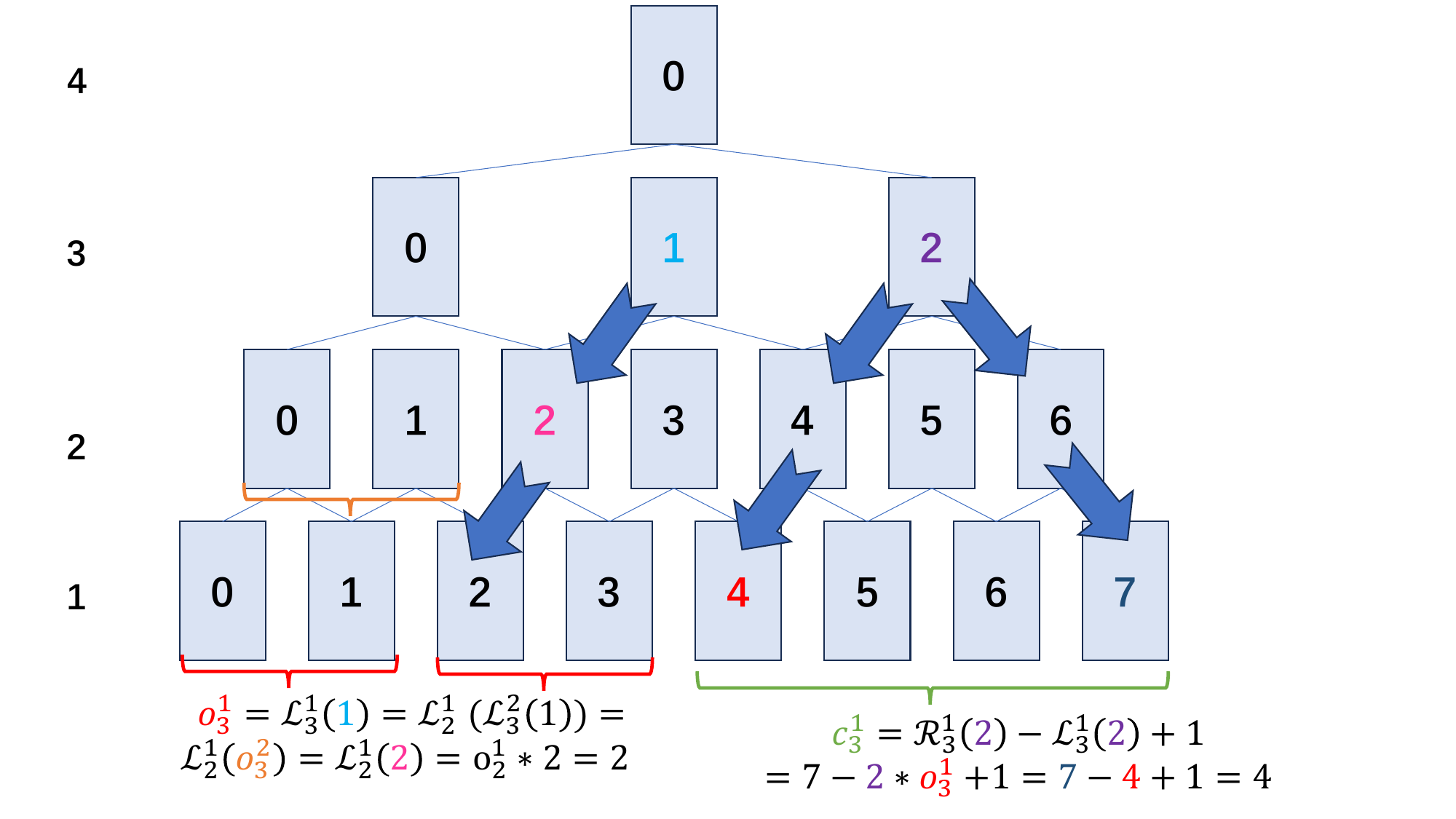}
\caption{Mathematical relations of ProPy. For each layer, indices start from $0$.  We focus on the prompt with index $1$ and $n_k-1$ for the $k$-th layer. The relations of its leftmost and rightmost descendant are transitive. }
\label{fig_algo}
\end{figure}

\section{Mask Construction Algorithm}\label{mask_algo}
The pyramid structure and attention masks are determined by the frame count $N_f$ and the structure hyperparameters $\mathcal{H}=\{(c_k,o_k)\}$ ($c_k,o_k$ are the number and offset of children). We provide one fast algorithm to construct the three attention masks $M^e_e$, $M^e_f$ and $M^e_v$ used for the Ancestor-Descendant Interaction Mechanism. There are two steps of the algorithm. 

\subsection*{Step 1: Cross-layer structure parameters}

First, we expand definitions in Equation (\ref{def_co}) to cross-layer structure parameters $c_{k_1}^{k_2}$ and $o_{k_1}^{k_2}$ between layer $\mathcal{L}_{k_1}$ and $\mathcal{L}_{k_2}$ ($k_1>k_2$):
\begin{equation}
    \begin{gathered}
    c_{k_1}^{k_2}= |\{ e^{k_2}_{j_2} | e^{k_2}_{j_2}\in \mathcal{D}(e^{k_1}_{j_1})\}| \\
    \mathscr{L}_{k_1}^{k_2}(j_1)= \underset{j_2}{\arg\min}\{e^{k_2}_{j_2} | e^{k_2}_{j_2}\in \mathcal{D}(e^{k_1}_{j_1})\} \\
    o_{k_1}^{k_2}= \mathscr{L}_{k_1}^{k_2}(j_1+1)- \mathscr{L}_{k_1}^{k_2}(j_1)
    \end{gathered}
\end{equation}

\begin{algorithm}[t!]
\SetAlgoLined 
\LinesNumbered
\KwIn{$N_f$, $\mathcal{H}=\{(c_k,o_k)\}$}
\KwOut{$\mathbf{H}$=\{$(k_1,k_2): (c_{k_1}^{k_2}, o_{k_1}^{k_2})$\}}
$\mathbf{L}=[N_f]$\ \ \ \tcp{length of each layer}
$K=len(H)$\;

\For{$k \gets 1$\  \textbf{to} $K$} {
    $\mathbf{H}[(k, k-1)] = (c_k, o_k)\;$\tcp{from $\mathcal{H}$}
    $n_k=(\mathbf{L}[k-1]-c_k)//o_k+1$\;
    $\mathbf{L}.append(n_k)$\;
}
\For{$k_1 \gets K$ \textbf{to} $1$}{
    \For{$k_2 \gets k_1-2$ \textbf{to} $0$}{
        $c_{k_1}^{k_2+1}, o_{k_1}^{k_2+1} = \mathbf{H}[(k_1, k_2+1)]$\;
        $c_{k_2+1}^{k_2}, o_{k_2+1}^{k_2} = \mathbf{H}[(k_2+1, k_2)]$\;
        $o_{k_1}^{k_2}=o_{k_1}^{k_2+1} * o_{k_2+1}^{k_2}$\;
        $c_{k_1}^{k_2}=\mathbf{L}[k_2]-o_{k_1}^{k_2} * (\mathbf{L}[k_1]-1)$\;
        $\mathbf{H}[(k_1,k_2)]=(c_{k_1}^{k_2},o_{k_1}^{k_2})$\;
    }
}

\Return{$\mathbf{H}, \mathbf{L}$}
\caption{Structure parameters}
\label{alg:H}
\end{algorithm}

We treat frame sequence as $\mathcal{L}_0$ with length $n_0=N_f$ for convenience. We calculate $c_{k_1}^{k_2}$ and $o_{k_1}^{k_2}$ with Algorithm \ref{alg:H}. The core codes are from line $12$ and $13$ that update $(c_{k_1}^{k_2}, o_{k_1}^{k_2})$ in an iterative manner. Here is a brief proof. We define an additional operation $\mathscr{R}_{k_1}^{k_2}(j)$ similar in  Equation (\ref{def_co}) to find the index of the rightmost descendant of $e^{k_1}_{j_1}$ in $\mathcal{L}_{k_2}$:
\begin{equation}
    \mathscr{R}_{k_1}^{k_2}(j)=\underset{j_2}{\arg\max}\{e^{k_2}_{j_2} | e^{k_2}_{j_2}\in \mathcal{D}(e^{k_1}_{j_1})\}
\end{equation}
It is not difficult to find relations:

\begin{equation}
    \begin{aligned}
        \mathscr{L}_{k_1}^{k_2}(j)=o_{k_1}^{k_2}\times j\\
        \mathscr{R}_{k_1}^{k_2}(j)-\mathscr{L}_{k_1}^{k_2}(j)=c_{k_1}^{k_2}-1
    \end{aligned}\label{relation}
\end{equation}

We obtain $\mathscr{L}_{k_1}^{k_2}(1)=o_{k_1}^{k_2}$ when $j$ is set to 1. Furthermore, $\mathscr{L}$ and $\mathscr{R}$ have a transitive property, as shown in Figure \ref{fig_algo}, which leads to: 
\begin{equation}
    \begin{aligned}
        o_{k_1}^{k_2}&=\mathscr{L}_{k_1}^{k_2}(1)
        =\mathscr{L}_{k_1-1}^{k_2}(\mathscr{L}_{k_1}^{k_1-1}(1))\\
        &=\mathscr{L}_{k_1-1}^{k_2}(o_{k_1}^{k_1-1})=o_{k_1-1}^{k_2}\times o_{k_1}^{k_1-1}\\&=...=\prod_{k=k_1}^{k_2+1} o_{k}^{k-1}
    \end{aligned}\label{eq_o}
\end{equation}

Similarly, we leverage the transitive property of $\mathscr{R}$ on events located at the rightmost position of each layer:

\begin{equation}
    \begin{aligned}
        \mathscr{R}_{k_1}^{k_2}(n_{k_1}-1)=\mathscr{R}_{k_1-1}^{k_2}(\mathscr{R}_{k_1}^{k_1-1}(n_{k_1}-1))\\=\mathscr{R}_{k_1-1}^{k_2}(n_{k_1-1}-1)=...=n_{k_2}-1    
    \end{aligned}
\end{equation}

Then, by applying equation (\ref{relation}), $c_{k_1}^{k_2}$ can be calculated as :
\begin{equation}
    \begin{aligned}
          c_{k_1}^{k_2}&=\mathscr{R}_{k_1}^{k_2}(n_{k_1}-1)-\mathscr{L}_{k_1}^{k_2}(n_{k_1}-1)+1  \\
          &=(n_{k_2}-1)-o_{k_1}^{k_2}\times (n_{k_1}-1) +1 \\
          &=n_{k_2}-o_{k_1}^{k_2}\times (n_{k_1}-1)
    \end{aligned}\label{eq_c}
\end{equation}

Equation (\ref{eq_o}) and (\ref{eq_c}) are implemented in an iterative manner in line 12, 13 of Algorithm \ref{alg:H}.

\subsection*{Step 2: Mask Construction}
Then we construct masks based on these structure parameters produced in Algorithm \ref{alg:construct}, filling attention areas with positive values layer-by-layer. For the sub-mask $M_{k_1}^{k_2}\in \mathbb{R}^{n_{k_1}\times n_{k_2}}$ from $\mathcal{L}_{k_1}$ to $\mathcal{L}_{k_2}$, ($k_1> k_2$), the attention scores are filled conditioning on relation $e^{k_2}_{j_2}\in \mathcal{D}(e^{k_1}_{j_1})$ and positions:
\begin{equation}
    M_{k_1}^{k_2}[j_1][j_2] = \begin{cases}
1 & \text{if } 0\leq j_2 -o_{k_1}^{k_2}\times j_1<  c_{k_1}^{k_2}\\
0 & \text{else}
\end{cases}
\end{equation}
Mask $M_{k}^{k}\in \mathbb{R}^{n_k\times n_k}$ on the same layer $\mathcal{L}_{k}$ is an identity matrix. $\widetilde{M}^e_f$ is constructed on the frame layer $\mathcal{L}_0$. $\widetilde{M}^e_v$ is same as the mask on $\mathcal{L}_1$.  $\widetilde{M}^e_f, \widetilde{M}^e_v$ are further expanded to ${M}^e_f, {M}^e_v$ with shapes $\mathbb{R}^{N_e\times (N_f\times N_s)}, \mathbb{R}^{N_e\times (n_1\times N_v)}$.

\begin{algorithm}[t!]
\SetAlgoLined 
\LinesNumbered
\KwIn{$\mathbf{H}$, $\mathbf{L}$}
\KwOut{$M^e_e$, $\widetilde{M}^e_f$, $\widetilde{M}^e_v$}
$N_e=\textbf{sum}(\mathbf{L}[1:])$\tcp{prompt number}
$n_1=\mathbf{L}[1]$\ \ \ \tcp{length of $\mathcal{L}_1$}
$N_f=\mathbf{L}[0]$\ \ \ \tcp{frame number}
$\mathbf{M}=\mathbf{zeros}(N_e+N_f, N_e+N_f)$\;

\For{$k_1 \gets K$ \textbf{to} $0$}{
    \For{$k_2 \gets k_1$ \textbf{to} $0$}{
    ${u}_{1}=\textbf{sum}(\mathbf{L}[k_1+1:])$\;
    ${v}_{1}=\textbf{sum}(\mathbf{L}[k_1:])$\;
    ${u}_{2}=\textbf{sum}(\mathbf{L}[k_2+1:])$\;
    ${v}_{2}=\textbf{sum}(\mathbf{L}[k_2:])$\;
    \If{$k_1=k_2$} { \tcp{same layer}
        $\mathbf{M}_{sub}=\mathbf{M}[{u}_{1}:{v}_{1},{u}_{2}:{v}_{2}]$\;
        $\mathbf{M}_{sub}.fill\_diagonal(1)$
    } \Else {
        $c_{k_1}^{k_2}, o_{k_1}^{k_2}=\mathbf{H}[(k_1,k_2)]$\;
        \For{$i \gets 0$ \textbf{to} $\mathbf{L}[k_1]-1$} {
            ${u}_{i}={u}_{2}+i*{o_{k_1}^{k_2}}$\;
            ${v}_{i}={u}_{i}+c_{k_1}^{k_2}$\;
            $\mathbf{M}[{u}_{1}+i][u_i:v_i]=1$\;
            \tcp{symmetrical}
            $\mathbf{M}[u_i:v_i][{u}_{1}+i]=1$\;
        }
    }
    }
}

$M^e_e=\mathbf{M}[:N_e][:N_e]$\ \ \tcp{$\mathbb{R}^{N_e\times N_e}$}
$\widetilde{M}^e_f=\mathbf{M}[:N_e][N_e:]$\ \ \tcp{$\mathbb{R}^{N_e\times N_f}$}
$\widetilde{M}^e_v=\mathbf{M}[:N_e][N_e-n_1:N_e]$\ \ \tcp{$\mathbb{R}^{N_e\times n_1}$}

\Return{$M^e_e, \widetilde{M}^e_f, \widetilde{M}^e_v$}
\caption{Mask Construction}
\label{alg:construct}
\end{algorithm}

\begin{figure*}[t!]
\centering
\includegraphics[width=2.0\columnwidth]{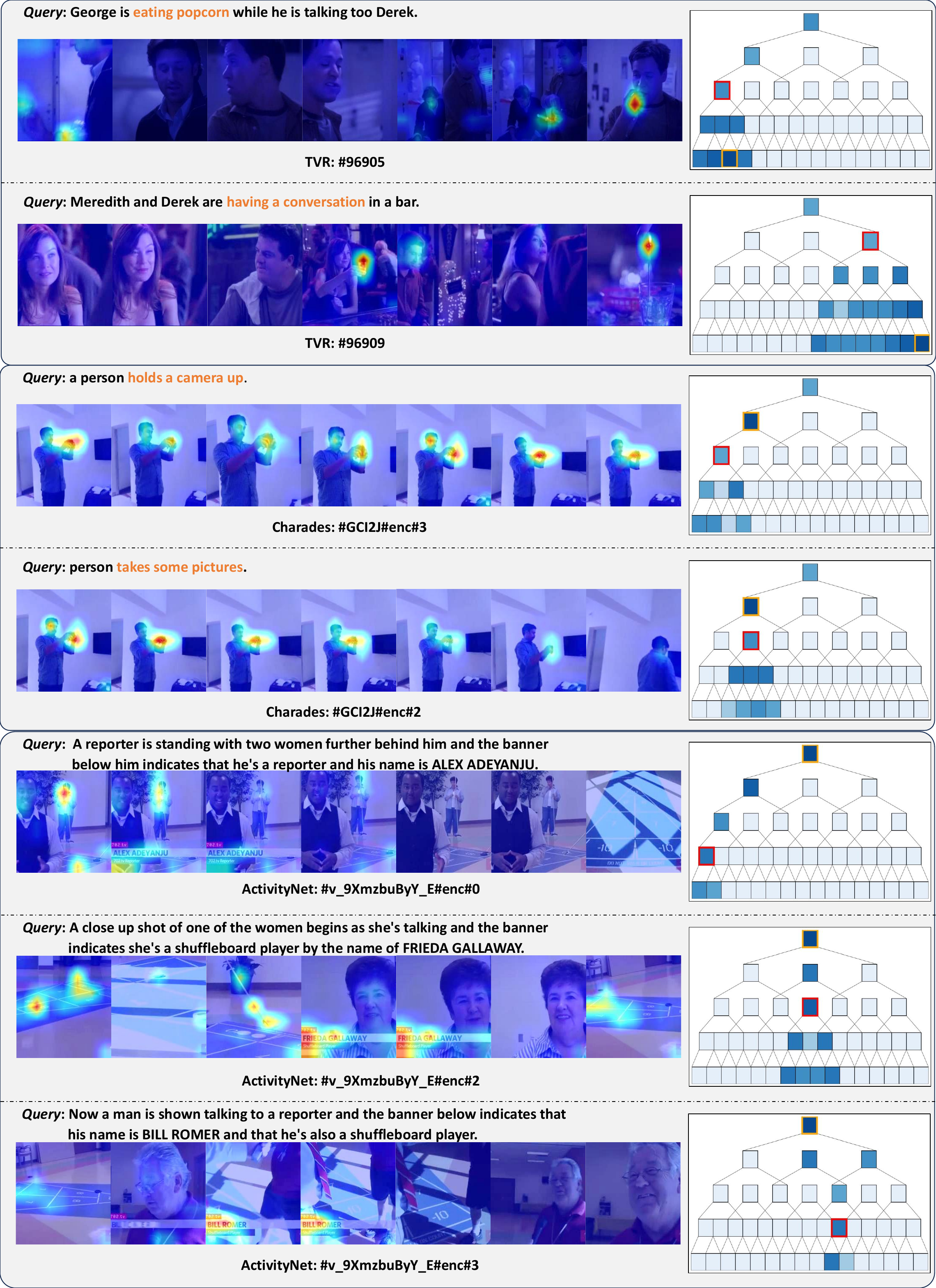}
\caption{More visualization results. Samples are selected based on the R@1 metric. Selected event prompts are highlighted with \textcolor{red}{red} borders. Events with the highest attention scores are marked by \textcolor{orange}{orange} borders.}
\label{fig_more_results}
\end{figure*}

\section{More Visualization}
We provide more visualization results in Figure \ref{fig_more_results} from TVR, Charades-STA and ActivityNet-Captions. It is notable that events from AcitivityNet tend to interact with high-level event prompts, like the global prompt. The reason is that many videos in ActivityNet are closely centered around a single theme, with longer  textual annotations and more complex dependencies. This forces prompts to interact with higher-level ancestors to obtain contextual information across a broader span.

\end{CJK}
\end{document}